# Video-Based Rendering Techniques: A Survey

**Rafael Kuffner dos Anjos · João Madeiras Pereira · José António Gaspar**

**Abstract** Three-dimensional reconstruction of events recorded on images has been a common challenge between computer vision and computer graphics for a long time. Estimating the real position of objects and surfaces using vision as an input is no trivial task and has been approached in several different ways. Although huge progress has been made so far, there are several open issues to which an answer is needed. The use of videos as an input for a rendering process (video-based rendering, VBR) is something that recently has been started to be looked upon and has added many other challenges and also solutions to the classical image-based rendering issue (IBR). This article presents the state of art on video-based rendering and image-based techniques that can be applied on this scenario, evaluating the open issues yet to be solved, indicating where future work should be focused.

**Keywords** Video-Based Rendering · Image processing · 3D Reconstruction · Free-viewpoint video · Data representation

## 1 Introduction

For a long time, video has been used in our daily lives as the media that more closely recreates an event as we live it in the real world. The recent popularization of personal video cameras and video content distribution has been pushing the scientific community to expand the traditional video format beyond its classical restrictions such as reproduction speed, which gave birth to slow motion videos, and most recently the viewpoint restriction.

Rafael Kuffner dos Anjos
Av. Prof. Doutor Cavaco Silva, 2744-016 Porto Salvo
E-mail: rafaelkuffner@gmail.com

João António Madeiras Pereira
Rua Alves Redol, 9 1000-029 Lisboa, Portugal
E-mail: jap@inesc-id.pt

José António Gaspar
Avenida Rovisco Pais, 1, 1049-001 Lisboa, Portugal
jag@isr.tecnico.ulisboa.pt

The process that uses video as input in order to create novel rendered content is generally defined as video-based rendering. This field shares goals and challenges with image-based rendering, while having the extra time dimension that is non-existent in its counterpart. By analyzing the visual content of these images, one tries to extract enough data to add processed information to the existing content or to create novel views that extrapolate the original experience.

The most popular goal on the field is the one of 3DTV (three-dimensional television). Not to be confused with stereoscopic displays that are commercially available nowadays, 3DTV aims to allow the user to navigate three-dimensionally around a point of interest, transforming the content to be correctly displayed from the new perspective. One early public example of the potential of this technique was the EyeVision[65] setup used on the Super Bowl XXXV. New viewpoints were not generated by this approach, but the user would get the natural feel of rotating around an object going from one camera view to another. Three-dimensional television has devel- oped [63] [50], but there is still work to be done towards achieving the final goal of having no positional restrictions.

The purpose of this article is to list and classify the different approaches on the VBR field according to a taxonomy described in Section 2, giving insight on their relationship. What are the lower-level requirements for higher level techniques and applications, and how can these techniques be expanded and developed to push applications into different objectives.

The outline of this document will start by defining video-based rendering and describing how we classify VBR techniques, followed by the state-of-the-art report on all the different processes on the VBR pipeline, comparing the most popular trends. Following, a conclusion and some insight will be given on what is the current trend of research, where research should be focused for the near future, and what is there to expect from future work.

## 2 Definition and Video-based rendering steps

Video-based rendering is a very broad term that has been applied to a wide range of techniques, sometimes in a broader way than it usually is, and other times focused on only a specific type of application. So it is important to establish the definition that will be used on this survey. The term was firstly used on the article by Schödl et al. [102] referring to image-based rendering techniques extrapolated to the temporal domain, using two-dimensional images of a scene to generate a three-dimensional model and render novel views of the scene.

The book from Magnor [84] defines video-based rendering as the process of fusing image-based rendering with motion capture in order to generate a novel view. A 3D reconstruction is not always needed to achieve the main goal of such applications; generating synthetic viewpoints. Borgo et al. [17] on their more broad survey classifies at a top level the techniques under the definition of video-based graphics (a more generalist definition for VBR), focused on creating new content (other videos or 3D reconstructions) based on video input, and video visualization that would encompass the attempts of allowing the user to see video from new/synthetic points of view not previously recorded.

The survey from Stokoya et. al [111] focused only on 3D time-varying reconstruction, more in line with the classical definition of Magnor [84], and would be only a subset of the previous classification, as also Szeliski [114] who stays with the classical definition.

We consider VBR to be the process of using two-dimensional time sequenced images to generate a new type of content. That excludes video editing or composing as VBR since the final result is of the same type as the input data. We choose to stay with the classical definition of VBR from Magnor, but not restraining it to generating 3D models out of video input, as the most recent trends of work on this area choose not to fully reconstruct a scene in order to generate novel views.

The definition of video-based rendering given above embraces a wide range of techniques and objectives, making a hierarchical classification of techniques for review purposes not viable. Few attempts of classifying VBR techniques as a whole have been made, commonly focusing on classifying each type of application or lower level techniques.

Although VBR techniques are highly varying on the required steps to achieve its objectives, one can assume a typical pipeline operations that are performed on a typical VBR process. Not all of the steps are essential depending on the application objective or input devices, and each one uses different techniques on the different steps.

This VBR pipeline is implicit across all surveyed works, and also reffered on the Survey from Borgo et al. [17]. On this survey we also consider the importance of data representation on the application level, which is a fac-

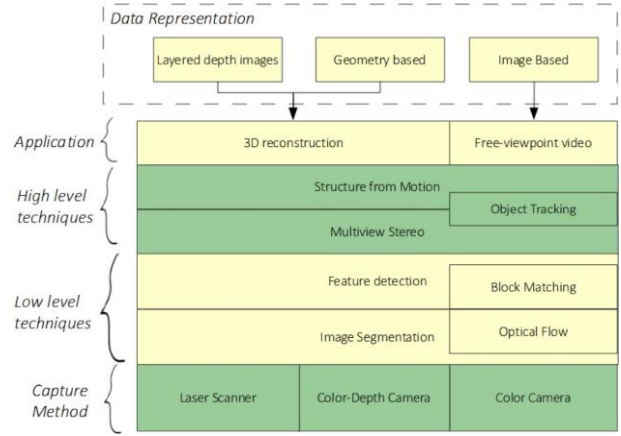

**Fig. 1** Diagram showing each level of the VBR pipeline from bottom to top and the reviewed group of techniques in this survey. Bigger groups mean techniques that are seen throughout all other types of techniques of different levels

tor sometimes overlooked on VBR processes and that becomes a bottleneck on high load scenarios. It is not added as a new step on the pipeline, but as something that is considered in this context. Figure 1 shows all of these steps, including each one of the subtopics that will be brought up on this survey. This pipeline consists of the following steps, bottom to up:

Capture: acquiring the data that will serve as input for the process, which can be performed using different setups and devices.

Low-level techniques: processing the raw input from each input source and adding low-level meta-information to these.

High-level techniques: combining images and/or low level meta-information in order to achieve a higher goal that is not yet a VBR application on itself.

Application: An application that can be used by a common user (e.g. Autodesk 123D Catch [57]) based on data using application-specific representation.

Other surveys have classified techniques according to taxonomies based on external aspects of the application such as level of automation, type of output and input information. We believe a review and description of related techniques for each step of the VBR pipeline, and a classification according to the use of such techniques brings more clarity to the inner workings of each application. We will classify techniques according to their choices at each level of the pipeline, that is the purpose of this review. Also, we want to highlight relationships between techniques on different levels of the pipeline, and evaluate the relevance of each. These conclusions will be presented on section 8.

## 3 Capture Hardware and Setup

Capturing images to be used in a VBR system implies choosing the input devices and the physical setup in accordance with the task at hand. Input data including or not depth, field of view, scene occlusions, calibration and synchronization are examples of aspects to consider for video-based rendering applications.

The main efforts in image and video-based applications are focused on capturing images with conventional color cameras [50] [52] [44] [24] [117], not only due to the lower cost of the devices, but the popularity of the developed methodologies (code publicly available) and the amount of data already available that could be used for applications such as showed on the work of Ballan et al. [8]. Besides being a bigger challenge than using more complex and informative data, it is of great interest to be able to use raw images for a VBR process.

Regarding the type of cameras suited for the VBR, there are some core requirements that should be met for the data to be useful. Data acquisition using all the cameras must be possible to trigger from an external switch so the several different sources can be synchronized, and the camera should be able to record in progressive scan mode, not interlaced half-images [84] which should be common in modern cameras. Camera resolution and recording speed are up to the application objectives, not being a general and essential parameter as the previous two. Other useful features are the ability to record raw pixel data, in order not to deal with images preprocessed by the camera internal hardware, flexible to high f-stop numbers, high dynamic range, good color properties, and other features. The book from Magnor [84] gives useful insight on some of the issues that should be considered for both image and video-based rendering.

The work from Goesele et al. [47] is an example of a mixed input that combines the raw images with an estimated bounding box for the object to be scanned. Also Ballan et. al [8] take other information as input such as available 3D models for a prior reconstruction of the scenery and better positioning of the cameras, since the input videos are not calibrated by default. The 3D model input does not always guarantee a better result, but having an initial geometry estimate does improve with the efficiency of the technique, as shown by the image-based rendering review from Shum and Kang [107].

Another input device that has been recently popularized is the Microsoft Kinect Sensor. It is currently the most popular depth camera on the market due to its real time nature and low-cost. Depth sensors were already an option on the past [34] but this device made it more accessible and complete with other built-in functions. Following the same trend, similar devices from Asus and most recently Intel were shipped with similar features due to the increasing interest on these devices. These devices use infrared structured light [5] [73] to generate depth maps of the scene that can be directly used to obtain a 3D point cloud representation of the scene. It has been applied for stereo view generation [5] using a single device as input. The work form Izadi et al. [58] also uses only one device as input, but captures a video of a static environment while moving the camera, similar to the work from Vogiatzis and Hernández [117] and also Zeng et al. [127], but the depth information from the kinect sensor allows them to reconstruct more than a single object and with better precision than the plain color images, due to a its precise measurement that is not set back by textureless regions as image-based stereo methods. This input method is not suited for outdoor scenarios, since its infrared is severely hindered by sunlight, washing out the projected pattern that needs to be recognized by the camera.

In a similar vein, Hornung et al. [54] describes octomaps as scene representation based in an alternative input that is the laser scan [59] [16]. Although laser scanners are still the best capture method when one intends to perform three-dimensional reconstruction of an outdoors static scene [69] [55] delivering results very close to reality, they are suited for static reconstruction, and thus fall off the classification of video-based rendering having significant dynamic content. Therefore they will not be compared side to side to the other input methods. There is although relevant and very successful work on the area of 3D reconstruction such as the city wide scan from [6] that used moving data captured from an aerial view with a laser scanner, and the 3D Michelangelo project by Levoy [74] that delivers ultra high quality reconstruction with a close sweep using a 3D scanner and cameras.

The new Kinect 2 device by Microsoft operates with the same technology as a laser scanner, and although it has not yet surpassed the ubiquity of the first Kinect in research projects, its real time capability of delivering realistic reconstructions is very promising on the VBR scenario. Also, the time of flight laser scanning technique of the new device works with phase-based encoding, which minimizes the multi-input interference problem.

Duan et al. [39] showed that is possible to perform fusion between depth maps from stereo cameras and Kinect sensors in real time, having an overall better result than using a single device. A similar attempt by Wang et al. [119] uses two depth sensors instead of just one, which gives a wider range of detection, but runs into the problem of mutual interference between the sensors. This is manageable in their specific setup, but displays one of the limitations of the multiple depth sensors approaches. One solution has been proposed by Butler et al. where a small motor is attached to a sensor, inducing a vibration that mitigates the conflict between different infrared cameras, since their lights will be in different frequencies due to the vibration.

When working with color cameras, multi input setups are more common and also necessary to perform stereo reconstruction. Apart from approaches similar to

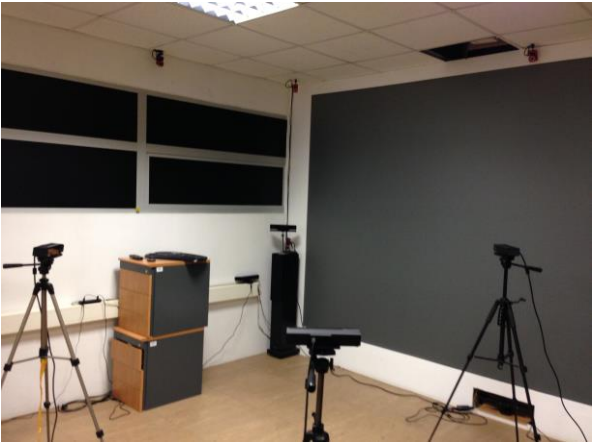

(a) Wide baseline of a low-cost setup for a 360 degrees capture of a subject.

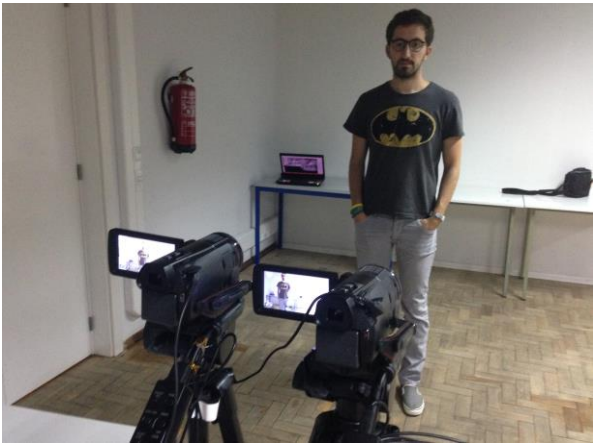

(b) Narrow setup with color cameras for stereo matching

**Fig. 2** Different capturing setups for VBR with different input devices.

Vogiatzis and Hernández [117] work, or image-based approaches [47] [131] [44] where still images from different positions are combined, one needs to have several synchronized inputs capturing the scene at the same time. They can be placed in a narrow or wide-baseline setup as seen on Figure 2. On the first, the cameras are placed closer to each other with little disparity between adjacent views, usually with each device parallel to each other. This is widely applied to Multi-view stereo works where feature based methods will have a better performance. The wide setup typically aims to capture a scene or object from all different perspectives, having the cameras placed further away from each other, where disparity between views is now desired, not avoided. To extract a visual hull or a silhouette during image segmentation, a wide setup is preferred.

On multi-streams approaches there is also the need of extrinsic calibration for the cameras, i.e. know the rel-ative positions between them, and also the distance to the objects in the scene. In controlled environments this can be done by using markers detected by the camera [112] [24], but on dynamic environments the most common approach is to track features using structure from motion [8] [58], providing a reliable position calibration for the camera. A parallel problem to this is the stream synchronization problem. This can be solved using an external trigger that starts all capturing devices at the same time, and one also tries to have identical cameras, and connected to a single computer that will pool the devices for images at a fixed rate. Again, this is only applied on controlled environments [24] [50]. In the case they are not connected to the same computer or the input streams are unrelated, as in the work from Ballan et al. [8], synchronization can be done by the audio track as done on this work, or manually if the result is not produced in real time [39].

### 3.1 Summary and Comparison

Currently the most popular input mechanisms for VBR are 2D cameras, given the fact that they are more accessible, and narrow-baseline scenarios usually need a big amount of input devices to produce the desired results [116]. For 3D reconstruction scenarios depth cameras such as the Kinect have been conquering their space since they can perform 3D reconstruction almost out of the box, but given the interference problems in using more than one sensor at the same time, they fall short on scenarios where a more complete view of a single object is desired and can not be captured by a single point of view.

None of the current popular input mechanisms can still rival the reconstruction precision of a laser scanner. The new Kinect 2 device will approximate that reconstruction precision with lesser cost and in real time, greatly enhancing the reconstruction quality on controlled scenarios. A short summary of each acquisition device strengths and drawbacks follow:

Color camera: Suited for any necessary setup, cheap, great amount of captured data available, main interest from other application fields. Works well on scenarios where reconstruction is not necessary, since performance on low textured scenarios is lower for 3D estimation.

Kinect-like depth sensor: Less expensive than laser depth sensors, reliable depth estimation on close distance/indoor scenarios. Not suitable for outdoor environments due to sunlight influence.

Laser scanner: Precise 3D data reconstruction, more expensive and slow capture mechanism classically associated with static environments. Now available at a faster frame-rate on the new Kinect two device.

Regarding multiple devices mounting (baseline), connecting (communication) and calibration of extrinsic and extrinsic parameters:

Calibration: Performed with markers on controlled environments, Structure from motion on uncontrolled scenarios.

Setup: Narrow baseline is used for higher detail and multiview stereo scenarios. Wide baseline for an application focused on a main object or performer, having a full view of it.

Synchronization: Crucial on video application, can be performed using audio cues, or done manually on controlled scenarios.

## 4 Low-level Techniques

This section describes the techniques that work directly on an unprocessed image or video with the purpose of attaching meaning to the analyzed content, working at the lowest level possible. Techniques such as Object detection that might use multiple low-level techniques will be described in Section 5. In this section we will approach optical flow estimation, feature detection and image segmentation.

### 4.1 Block Matching

Techniques in this category are one of the classical ways of extracting information from moving pictures, these algorithms consist in dividing a frame into blocks of size N x N, and matching them against equal sized candidate blocks on an area around the original location. After finding the best matched block, the difference (Motion Vector) between time t and t+1 is recorded.

Block matching is mostly used in video compression to remove temporal redundancy within frames, detecting what are the actual differences that need to be transmitted in order to fully reproduce the video. Although block matching is not commonly associated or applied to video-based rendering, it can be relevant on scenarios where video or data compression is desired.

Full-Search Block Matching Algorithms (FSBMA) provide the optimum solution to the problem, but given the high computational load they require, several fast BMA's have been developed assuming different simplifications that provide a lossy solution. The reference algorithm for the implementation of MPEG4 [130] applies a search strategy called Diamond search that restricts the search space using a diamond shaped pattern. The results do not equal a FSBMA, but greatly outperform its prior strategies such as cross search [46] and three-step search [67]. Other simplifications can be made in order to make the BMA more efficient such as simplifying the matching criterion, bit width reduction and predictive or hierarchical search. The survey from Huang et al. [56] describes these approaches in detail and provide in depth comparison between them. A more recent comparison can be found on the article by Yaakob et al. [122].

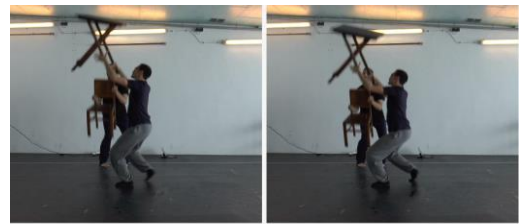
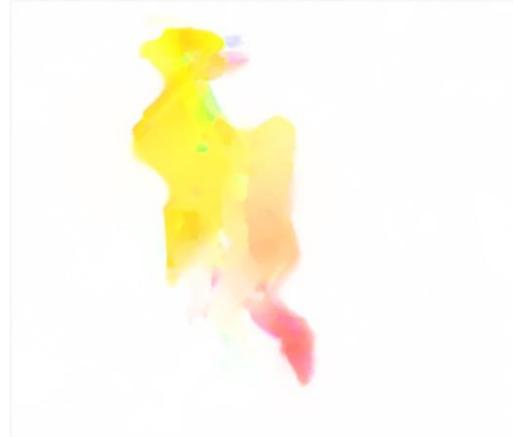

(a) Output of a flow analysis algorithm [121]

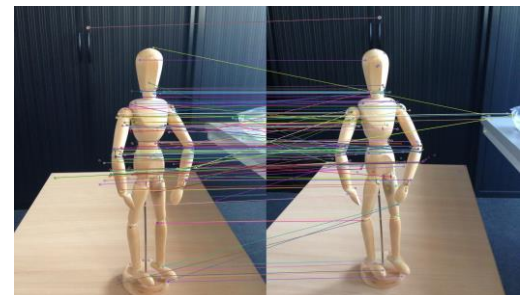

(b) Feature detection output using SURF [10]

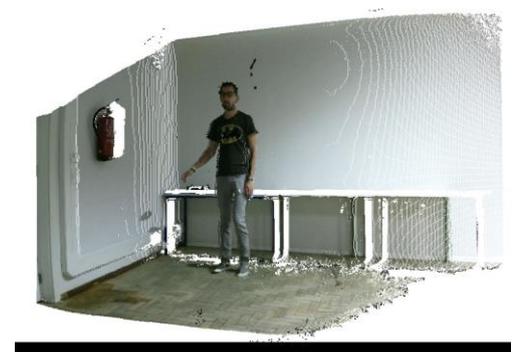
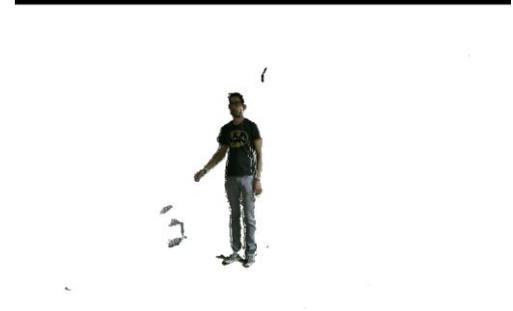

(c) Background removal on a point cloud dataset

**Fig. 3** Example outputs of different low level techniques. All of them add some processed information over raw color data.

The class of algorithms that come closer to the correctness of FSBMA are the Fast Full-Search Block Matching Algorithms (FFSBMA). The main idea behind these techniques is to perform a simple computation to determine whether a candidate block can be the optimal one or not, in order to avoid doing unnecessary computations. Successive elimination Algorithm [77] is a good example, where a faster calculation is applied to the block and compared to the up-to-date minimum. If his value already exceeds the minimum, further calculations are avoided by skipping this candidate. Although the results are very close to FSBMA, they are not always equal. When two candidates have the same estimated value, it depends on the search order to decide which one will be discarded. These small differences do not compromise the final result in a noticeable amount.

Cuevas et al. have used a Differential Evolution approach [32] to reduce the number of search locations, using a fitness function to decide which regions are worth to be searched. A more recent work [33] poses the problem as an optimization task for the Motion Vector and uses an ABC (Artificial Bee Colony) algorithm to achieve a fast answer without compromising the result. They claim their approach does not easily get trapped in local minima due to the non-linear characteristic of the ABC operators, and that the efficiency is comparable to the fastest algorithms.

## 4.2 Image segmentation

Often on Image processing applications, there is the need to identify and group pixels that share a certain meaning such as belonging to the same object, image region, or a certain color. This process of image segmentation is one with a generic purpose, that must be fine tuned and trained to solve each desired problem. On the field of Image based rendering, it is applied mainly to distinguish the applications point of focus from the background, which would introduce noise to the process. A general review on Image segmentation methods is given by Pal and Pal [93], and foreground detection by Bouwmans et al.[18], while we will focus mainly on background detection techniques, since they were the only class applied on image or video-based rendering applications on the reviewed literature.

The simplest method for image segmentation is Chroma-key [12], where the background is already colored in a specific color before previously to the recording, being easily identified after the images have been captured by checking the pixels with a similar color to the one used to identify the background. On applications where the objective is not to use real life data for IBR or VBR, but analyzing a specific object in a controlled environment, chroma key can be the best option. The work from Liu et al. [80], and also the work from Hauswiesner et al. [52], use Chroma-key to subtract the background and obtain silhouette information to perform free viewpoint videos (FVV) of human beings. On such applications where background and foreground distinction is needed for the application objective, this is a viable alternative. Ning et al. [91] also depend on some sort of user input, but simpler than a chroma key background. On a first step a simple segmentation process such as mean-shift [30] is used, then user input is requested to insert two distinct simple markers on a foreground object and on a background object. Regions containing each of the distinct markers are tagged as being foreground and background, and then a merging process is performed to achieve the real desired segmentation.

An interesting and very efficient approach to the problem was given by Arbelaez et al. [4], who try first to solve the problem of contour detection, and uses its output to perform image segmentation using the lines as guidelines to separate the image into different regions, achieving surprisingly good results that closely match the human segmentation. This approach is similar to the reference work from Del Bue and Agapito [36] on grayscale images using contour detection as visual cues to segmentation. Another unsupervised method for images was presented by Joulin et al. [60], that performs background and foreground detection using only images from the same object in different situations, this is achieved using SIFT features [81] and clustering techniques, putting the task of image segmentation as a high level task that does not simply use images as input anymore. Galun et al. [45] and more recent work by Alpert et al. [3] apply a bottom-up approach where pixels are grouped considering not only intensity values, but likelihood of having the same texture. The second approach uses a probabilistic algorithm that requires less manual input and has comparable results to the first.

The most popular class of algorithms that can be applied to video to perform segmentation is change detection, since from frame to frame the background will most likely stay the same, it is easily detected by these techniques. The simple yet popular approach of image differentiation [100] works by simply thresholding the difference between two images at a global level. An example can be seen on figure 3(c). There are also more complex techniques such as Change vector analysis [23] and Image ratioing [37]. Although popular, these techniques suffer with noise and illumination variation [22].

Other approaches use a statistical hypothesis test instead of a simple value comparison. Grouping the pixels into blocks, the hypothesis of change is tested using likelihood ratio or significance tests, or other probabilistic models. When an image sequence is available such as in a video, predictive models can be applied, exploring the relationship between space and time on neighborhood of pixels. A more detailed description and several references for each class can be seen on the article by Radke et al. [99].

## 4.3 Feature extraction

Feature extraction is a classical low-level image processing challenge that is applied to several applications, such as object recognition, tracking or image and video-based rendering. Global features analyze the image as a whole and provide us with information on a general level, such as illumination and color. Since they are unable to provide us with precise object information, their use in Video-based rendering is limited. Therefore we will be focusing our description on Local feature detectors and descriptors. The main use of these techniques in video or image based rendering is to detect specific points of an object of interest, in order for it to be identified on a different image or on a future frame of a video sequence.

Feature detectors usually look for patches with image points with high spatial directional derivatives i. e. the so called image corners. A derivatives or, more generally, a saliency map is created, and the local maxima are selected as the features using the squared gradients matrix. The approach from Harris and Stephens [51] uses a corner and edge detector based on a local correlation function to detect these features. Lucas et al. [82] analyze a window around each point to perform similar calculations, trying to maximize the determinant of such matrix. By doing so, a high variability on x and y can be detected and assigned as a feature. Following work from Lindberg [79] has a different approach, using the difference of gaussians to have features that can be detected and matched between objects at different scales on different images.

After detecting key features, descriptors are created to characterize the image around the location pointed by a detector. Scale Invariant Feature Transforms [81], or SIFT, are the most popular ones, calculating the descriptor on different scales and rotations, being able to be matched against a large database of images of the same object in different scales and positions. The shape context descriptors [13] are a similar approach but based on edges. Close on popularity are the Speeded-Up Robust Features (SURF) [10] (Figure 3(b)) that represent features using sums of Haar wavelet components, that represent more of the global image intensity, in opposition to the more local SIFT features, and are calculated a lot faster with comparable precision.

An extension for SIFT applied for face meshes resulting from face scans has been developed by Maes et al.[83], showing how well and adaptive SIFT descriptors can be. Another example is the work from Goesele et al. [72] where SIFT features are used in the context of creating global features for spatial recognition. A more detailed view on detectors and descriptors for image-based applications, refer to the survey from Moreels and Perona [89]. Boyer et al. [19] perform a comparison between different approaches putting a human shape through different modifications such as scaling, sampling, holes, or affine transforms, where different techniques and their precision in such situations are analyzed.

Specifically for videos, some approaches extend the SIFT idea to adding a temporal dimension to it. It has been used by Scovanner et al. [103] to perform action recognition, using a sequence of sift descriptors to represent each action. It is a similar approach to a motion field, using gradients over time to assign an orientation to each feature, and by combining them into a vector a temporal feature is created. The approach from Ben Ahmad et al. [14] accomplishes a similar goal using an accordion representation, that tries to place pixels with high temporal correlation in adjacent spaces. Moving points are detected after background removal and motion detection, so it is better suited for single object tracking. The recent work from Tang et al. [115] on Bold Features is also well suited for video scenarios where light variance and blurred images are a common issue. On these specific scenarios its performance was better than SIFT or SURF features.

## 4.4 Optical Flow estimation

Optical flow is a velocity field in the image which transforms one image into the next image in a sequence [53], one of the few low level techniques that is exclusive to video (Figure 3(a)). While the other techniques will mainly examine a still frame, this class compares consecutive frames in the flow of time to extract information on how the objects captured in video move on time. In this section, differential methods (local and global) are discussed. For a more detailed discussion including energy-based methods, phase-based techniques, and region-based methods, please refer to the survey from Barron et. al [9] or the more recent technique comparison by Del Bue and Agapito [36]. For an overview on Motion Estimation algorithms the recent study from Phillip et al. [95] explains on higher detail the basics of these techniques.

Differential methods are the most common techniques to do flow estimation. Assuming only small changes between consecutive frames, spatio-temporal derivatives of the image intensity or other measures are calculated between frames, estimating the optical flow as velocity vectors. Differential methods can be roughly classified as local or global.

Global methods use a smoothness operator or a regularization factor applied to the whole image in order to avoid the aperture problem [15] of local methods, where motion information can not be estimated since the indicators of motion of the analyzed object are outside the area looked upon. This problem is normally noticed on large motions that are covered in the whole picture. The downside of these operators is that since they are applied to the image as a whole, they are more sensitive to noise, propagating errors through the whole image [21]. The most referred global method is the work by Horn and Schunck [53] , where the authors claim the apparent velocity of the brightness pattern varies smoothly through the image. Therefore a measure on the global brightness of the image is used as a second measure to enforce the local calculations done with first-order Taylor expansion of image motion.

Local methods suffer from the aperture problem, but they avoid the error propagation of global methods as they only calculate the motion flow using local information from the pixel neighborhood of the patch currently being evaluated. The most popular local method is the work from Lucas et al. [82], where the spatial intensity gradient of the images is used to find a match in the following frames using Newton Raphson iteration. The result is a more robust but less dense flow field. More recent implementations such as the work from Bruhn et al. [21] attempt to use the best of both algorithms. By cumulatively adding the local contributions to the histogram, the regions with lowest contribution are detected as the ones with highest confidence. This measure is then applied to the global smoothening function, that will have a higher effect on these areas. Also hybrid approach is the one from Chen [29] that combines differential methods with block matching techniques, since each one of them has better performance in different situations. With the combined efforts they can achieve a precision comparable to the full search algorithm, with better efficiency.

### 4.5 Summary and Comparison

The four techniques serve different purposes by themselves, but all operate at the same level of the pipeline. Block matching has not been applied in many applications related to this topic, being mainly used for video compression and analysis. Optical flow is one of the most promising and useful technique for video-based rendering since it is the only one that takes advantage of the unique time component of videos, but has not been used in many higher level techniques to this day. It is used as essential information for object tracking [124].

Image segmentation plays a major role in several applications since usually the focus of the applications is to enhance the visualization of a single object that can be considered in the foreground [8] [24] [80]. It has also been successfully applied to video taking advantage of the time component [100] and not only analyzing still frames. As with Optical flow, this component can be explored to a bigger depth in video-based applications.

Feature detection can be considered the most important low level technique given the range of different uses its output can have. They can be used for recognition, reconstruction, analysis, and other types of applications. The fact that there are known effective implementations available to be used enables researchers to take them for granted when used on a general scenario, that no special features need to be tailored to fit its purposes, such as object tracking applications.

Regarding their relation with the capture methods, feature detection can be used to describe three-dimensional data [76], and also segmentation can use geometry information to achieve more precise results [76]. Optical flow and block matching have not been explored on the Color-depth scenario due to them being exclusively applied on video, and an efficient specification and format for three-dimensional video having not yet been developed. Following is a short summary of strengths and relation to higher or lower levels of the VBR pipeline, regarding the covered techniques on this section:

Image Segmentation: Essential technique for applications with a human focus on the first plane. Has different algorithms and solutions that must cater to the specific capturing scenario. Can be performed using any type of input.

Block Matching: Typically used for image compression [130] of videos coming from usual cameras. Not commonly applied on VBR yet.

Feature Detection: The most important low level technique, required by all of the higher level techniques on VBR. Different types of features exist, each one of them performing better on specific capture devices, layouts, data quality, or situations.

Optical flow: Used extensively by object recognition techniques, works with color cameras. Several algorithms developed with a varying compromise between speed and efficiency.

## 5 High-level Techniques

After obtaining features, image segments, or optical flow, this information can be used as input for a new processing step that outputs even higher level of information about the capture, without creating a user-targeted application. We call these steps High-level techniques. Such applications which are result of applying these methods will be discussed on section 6, together with the relations between high level techniques used to produce a more concrete result.

### 5.1 Multi-view Stereo and Photo-consistency reconstruction

3D-reconstruction is a classical challenge for computer vision and image-based rendering. Multi-view stereo (MVS) has been the most standard approach [126] [92] [43] [40] [104] [96] [11] [20] , seeking to combine images captured from known positions to reconstruct a complete 3D model of a given scene. In order to achieve this objective, lower level techniques are used to properly identify and match image features that enable a full reconstruction of the scene using triangulation on the optical rays.

One group of techniques that was initially an alternative to the matching features approach [86], but recently has been used coupled with MVS is the photo-consistency based reconstruction that can be applied when a rough estimate of the input model or scene is provided [70] . These approaches tend to perform poorly when facing occluded areas, but when combined with MVS can surpass these issues. Our review focus on most recent and popularly used algorithms, describing them on a spectrum of solely multi-view stereo, combined techniques, and only photo-consistency reconstruction.

The work from Furukawa and Ponce [44] is a recent example of classical Multi-view Stereo. Using difference of Gaussians and Harris operators, an initial representation of patches is created and expanded iteratively until a dense reconstruction is achieved, and then filtered using visibility constraints. The principle behind the approach is similar to pioneer works such as Okutomi and Kanade [92], but introduces a novel patch-based representation that suits very well the reconstruction task.

Vu et al. [118] start by using feature detectors and triangulation to create an initial point cloud and then estimate depth maps from pairs of images. A surface is estimated after outlier removal, using optimization algorithms with photo-consistency measurements, and regularization to attenuate the noise of the point cloud. This work shows how both techniques have been combined, and how these hybrid solutions can solve problems that a more classical approach would face. Fan and Ferrie [42] apply a similar approximation that uses the photo hull as an estimate for the first step. This estimate is the basis for regularization of the standard multi-view stereo, in contrast to the approach from Vu et al. [118] where the photo-consistency measurements are only applied after the MVS process.

Esteban and Schmitt [41] and also [68] start with a depth map computation from stereo paired images, and fuse them into a single volume to represent the desired model using silhouette image information to validate the generated depth maps.Work [41] is based in the snake framework for optimization, while [68] applies graph-cuts.

The recent work from Lafarge et al. [71] proposes an iterative improvement approach, representing a classical photo-consistency reconstruction approach. The starting point is a rough estimate of the 3D model that one wants to reconstruct, and interactively it is segmented and the extracted primitives are back-projected into the input images to apply photo-consistency measurements. The process is repeated until no improvements can be made, and the back projection closely resembles the input images.

On video sequences, Liu et al. [80] has successfully reconstructed human performers captured on a wide baseline setup against a green screen using a similar approach to the one of Vu et al. [118] i.e. creating an initial point cloud with feature detection and epipolar geometry, and refining it with the visual hull easily extracted from the background segmentation with chroma key. Another approach that has been tested on video sequences is the one from Zitnick and Kang [131] that uses image over-segmentation and color-based matching. This approach was shown to be more robust to noise than feature tracking.

These works state a strong point that classical MVS can be applied to videos with effectiveness and deliver good results. For a different insight and classification on MVS, the survey from Seitz et al. [105] classifies the different techniques according to aspects as the scene representation, photo-consistency measure, visibility model, shape prior, reconstruction algorithm and initialization requirements.

## 5.2 Structure from Motion

Structure from motion (SfM) is the extraction of a 3D structure from a set of 2D sequence images based on the motion of the structure. It normally works under the assumption of having a static environment, so the motion is either a result of camera movement, or the environment as a whole rigid body.

The work from Chen and Pinz [27] performs the classical SfM approach of feature detection and position tracking, being able to recover a sparse 3D structure from an image sequence after estimating the camera parameters. The more recent approach from Crandall et al. [31] has also applied the principle of Structure from Motion to sets of uncalibrated images instead of a video sequence, achieving 3D reconstruction and camera parameters using a belief propagation framework for their optimization algorithm.

The process of SfM in some scenarios, such as the work from Vogiatzis and Hernndez [117], can be classified as a type of multi-view stereo, since one can consider the sequential images of a still object as different images from different points of view. MVS and SfM are closely related. SfM being a less costly algorithm can be used to extract a rough estimate of scene structure before a more costly reconstruction is performed using MVS on this particular work. As seen in Section 5.1, MVS can be applied to video input for 3D reconstruction. SfM more recently tends to be applied in conjunction with MVS.

The whole city reconstruction methodology from Agarwal et al. [1] takes such an approach. It is able to reconstruct the city of Rome from a set of input images obtained from Flickr. Similarly, on the works of Ballan et al. [8] and Snavely et al. [109], SfM is used solely to estimate camera parameters and the general structure as the first step for the application.

### 5.3 Object Recognition and Tracking

Object recognition and tracking are methodological components found in many computer vision and VBR applications. The ability of identifying and tracking a scene element in a video is advantageous in scenarios such as motion and gestures recognition, automated security, AI navigation on the real world, and many others. The common approach to object detection is based in using a set of object features to identify the object in different scenarios and then use those identifications as training data for a classifier such as a decision tree, neural network or support vector machines.

Relating to the techniques mentioned in this chapter, the key element on a recognition and tracking algorithm is the choice of the tracked features. Scale invariant features [81], motion flow [130], and object shape or silhouette [13] [18] are some of the common elements to be detected and tracked on a video sequence. The surveys from Yilmaz et al. [124] and Prasad [98] give an insightful look on these techniques classifying them accordingly.

When applied to a known type of object to which the classifier is trained such as facial recognition or gestures, object recognition and tracking methodologies are already part of our real life and present in our mobile phones [101] [113]. An example is the Microsoft Kinect [88] device that recognizes several concurrent players in front of its depth and color cameras. Never the less challenges remain on untrained scenarios.

### 5.4 Summary and Comparison

While Multi-view Stereo is classically applied to still images, it is still the most used high level technique given the large background research on still images providing solution for different scenarios even on video-based apps. Since, as discussed in Section 5.1, they can be as effective on those scenarios. More recently MVS has been combined with Structure from Motion on video-based scenarios for validation [41] and also [68], noise removal [118] and efficiency [42].

Structure from motion takes advantage of the time dimension that is not present in still images, but it is not as effective on reconstruction as MVS, being a faster but less precise technique. It is usually applied to camera calibration and parameters estimation. Although a full reconstruction algorithm using SfM would not be as precise as MVS, its usage can be expanded for video reconstruction scenarios where moving objects need to be reconstructed.

Object Recognition and Tracking can be a standalone application on its own, but can also provide important meta-information for FVV applications or even Reconstruction tasks where a single object on the scene is the center of attention [8] and one needs to keep track of its position. Following is a summary of the main uses for each technique. Being naturally different and more complementary than concurrent, there are few points to be compared between each of them.

Multiview Stereo: 3D reconstruction on images or video, stereoscopy synthetization. Most common high level technique on VBR.

Structure from Motion: Camera parameters and motion estimation for uncalibrated video sequences. Rough estimate of 3D structure based on color images.

Object Tracking: Mostly necessary on interactive VBR applications, or scenarios where there are specific points of interest such as sports or artistic performances.

## 6 Application Objectives

The last classification we can make on VBR is regarding the final purpose of the application, from a user perspective. The survey from Borgo et all. [17] classifies it in two main different goals: Video visualization and video-based graphics. We will focus more on the second type of applications that aim to create novel content based on video, while the first one focuses more on enhancing a video presentation, that is not the focus of this chapter. While there might be other types of specific applications, we focused on the most prominent types of work on the community; Free viewpoint video and 3D Reconstruction.

### 6.1 Free-viewpoint video and 3D-TV

One of the first big motivations for video-based rendering is enabling the idea of a Free viewpoint-TV (FTV), where the spectator takes a more active role and is able to choose the point of view from where he wants to watch the said broadcast. This is the most common form of FVV and what is commonly targeted commercially. A reference example is the work from Kanade [61] that was in the coverage of the Super bowl XXXV, where not the users, but the broadcasting team was able to cycle seamlessly around the several cameras set up in the stadium to give more insightful replays. A similar recent product by Vizrt [78] has been extending the functionality to several other sports adding other overlays and video processing effects to give enhanced replays, such as player and game elements tracking.

The article from Goorts et al. [50] explains a similar setup used on a football event. Using a semi-wide baseline setup the images are captured, and cameras are calibrated using SIFT features matching on a preprocessing step. At runtime image segmentation is performed using some of the calibration images to identify the background. Novel viewpoints are then computed by estimating a depth map for the new chosen position to render the background, and the players are placed accordingly to the detected depth.

Ballan et al. [8] has produced a similar result but on the more challenging scenario of casual and uncalibrated video recordings. Unlike the previously mentioned setups, the videos have no relation with each other, so additional steps of video synchronization and camera parameters calibration need to be performed. Background is reconstructed using structure from motion with the videos and also images or 3D models of the captured scene, and camera positions are then estimated relative to the sparsely reconstructed background model. Segmentation is performed with user input help, given the unstable nature of the video recordings. After the foreground object is painted in a couple of images for each video as training data, the rest of the process is performed automatically with object recognition techniques. On runtime, the user can choose to change between the viewpoints of different cameras with natural transitions around the foreground object.

A closely related topic to FTV is 3D Television, where the focus is on displaying a stereoscopic image to the viewer. That can be performed by synthesizing the views for each one of the eyes using similar techniques to the ones used on FTV, meaning that a FTV system that produces synthetic views can also deliver stereoscopic vision to its users depending on the used display. The survey from Zhang [128] provides deeper insight on this matter, showing where the methods overlap and differ. Tanimoto [116] proposes ray-based FTV system that enables them to perform early attempts in having something similar to an holographic display, using a very complex narrow baseline setup for capturing. The author proposed his system as the MPEG standard for FTV, bringing up the question about distribution of this content, and that is where block matching and motion flow algorithms can play a big part, since they are one of the main components of 2D video compressing and encoding. For 3D video, no definite answer has yet been achieved. The work from Yang et al. [123] and also Do et al. [38] already use a wider baseline setup using also a depth sensor, and the second [38] already considers block matching techniques for inpainting, which is a step on the expected direction of using lower level video techniques for VBR.

## 6.2 3D Reconstruction

Unlike FTV where only the general structure of the scene is necessary to estimate the positions of some elements to be rendered in a synthesized point of view, these applications aim for a full 3D reconstruction of the environment or object in the form of a mesh or point cloud that resembles the original as best as possible.

The works mentioned on Sections 5.1 and 5.2 mostly aim for 3D reconstruction [1] [80] [44]. Multi-view stereo is one of the main approaches to reconstruct 3D models based on real world captured images since image-based rendering started to be researched. While 3DTV/FTV is more closely associated to video-based rendering, 3D reconstruction usually works with still images [44] [41], since the focus is normally to reconstruct on maximum detail a single object captured by different cameras. It is also applied combined with other techniques for more detailed reconstruction such as seen in the work from Beeler et al. [11] and Bradley et al. [20].

Recently initiatives to push 3D image-based reconstruction as a consumer app on a mobile phone have been done. Autodesk 123D Catch [57] allows users to perform this task based on sequences of pictures taken of an object and sends it to a remote server to process and generate a textured 3D model. Google's Project Tango [49] aims to deliver a spatial reconstruction based on video captured on a navigating mobile phone using the device's accelerometer to detect positional changes and fully reconstruct an environment.

Few attempts have been made to fully reconstruct in detail a moving scene or object, having the work from Liu et al [80] as one of the closest to achieve this specific goal, but was only applied to a single performer on a controlled environment. Also a similar goal was shared by Stoll et al. [110] where an articulated character was reconstructed based on video footage of a performer in a controlled environment. The follow up work on Kinect Fusion from Zeng et al. [127] reconstructs a static environment form a dynamic scene in real time, allowing a simplified visualization of the dynamic elements that are left out. [120] uses shading cues and silhouette cues to estimate the pose and shape of performers, reconstructing them fitting the depth data to human meshes. On a similar related work by Li et al.[75] the same technique is performed but also allowing the human reconstructed actors to be re-lit by new settings.

There are several challenges in performing a full 3D reconstruction of an environment, some of them being the precision of the reconstruction itself, that can be highly compromised by a dynamic and moving environment with occlusions between scene elements. But also if aiming for a FTV approach, a full 3D reconstruction might not provide the best of results when comparing to a synthesized interpolated view as described on Section 6.1. But on a different scenario where the goal of the application can be something different, this goal might be more relevant. We recommend the survey from Stoykova et al. [111] can give some more insight on this area under a different point of view.

## 6.3 Summary and Comparison

Free viewpoint Video is without a doubt the most promising output of a VBR technique, but it still slightly limited by the available output displays [116] and typical user interaction paradigm with these. On a virtual environment though, this technique can be very useful and user controlled videos can be a reality in the near future.

3D reconstruction for static environments has greatly advanced and shown outstanding results even on outdoor scenarios [1], the current focus is on articulated/flexible objects and dynamic scenarios, where mutual occlusions between objects still pose a great challenge.

Although the objectives are by definition different, 3D reconstruction can also be used to enable the user to relive a recorded scene from a different point of view [24], but the main focus on 3D reconstruction has been on static environments where results are guaranteed to be better. Also, for Free viewpoint video, view interpolation shows to have a better result than a full reconstruction with the currently developed techniques.

Several VBR recent works make use of a sparse 3D reconstruction to achieve their objectives [8] [50] [78], but performing a full reconstruction is currently not viable due to the amount of uncompressed data created being not manageable for a rendering task. Fully reconstructing an environment would not only be useful to perform Free-viewpoint video, but also different applications that would use this type of media, since more complete environment information would be available.

## 7 Data representation

Although this is not a defined step for the VBR process as the previous sections (see section 2), inefficient data representation can be a bottleneck on VBR techniques. This step is sometimes overlooked on VBR applications that focus on other aspects, but it can play an important role on scenarios with a high amount of data, or real time requirements. Inefficient data representation and structure might increase the necessity of disk lookups or complex real time calculations, which might hinder us from reaching these objectives.

Simple classical representations are Meshes and Point clouds [66], which can be created not only by IBR processes, but also by artists or other capturing devices. Being classical representations for 3D data, they are supported by default on the traditional rendering systems. Each one of them have different formats to be stored and compression issues related to them. They are still suited to Image and video-based rendering [109] [110] but are not specially tailored to VBR. Set of patches [94] have been used instead of traditional point clouds in order to provide a better visualization for this type of data, but are still better suited for a static visualization, and not the type of VBR applications described in this survey.

The View interpolation process described by Chen and Williams [28] uses arrays of images as their representations and are suited for IBR and VBR systems that use color cameras as their input devices, and do not perform three-dimensional reconstruction. This is the process that is nowadays most commonly applied on VBR systems [8] [50], but is suitable only for view interpolation, not for more complex restorations tasks. This rep-resentation is on the opposite end of the spectrum when compared to the previously addressed. Here we do not have any three-dimensional information but the cameras location, while on the previous one we had only the three-dimensional information of the scene.

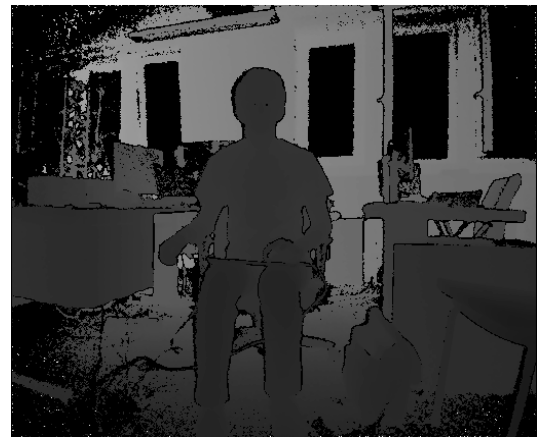

(a) Depth map

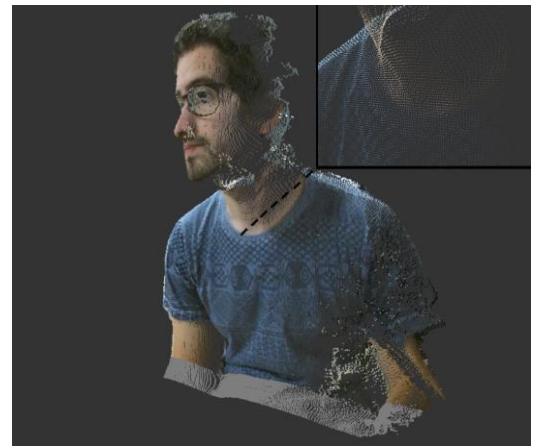

(b) Point Cloud

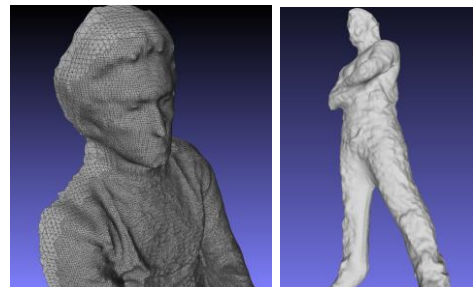

(c) 3D mesh      (d) Visual Hull

**Fig. 4** Different data representations for a captured human being.

One alternative representation is the visual hull [86] that are well suited for wide-baseline scenarios where one object is the purpose of visualization. Using the perceived silhouettes on the different viewpoints a bounding box for the object is defined with mapped textures onto it. Figure 4 shows an example created for a standing person. This representation has the limitation of not being able to represent concave surface regions. It is visually similar to a mesh.

The Layered Depth Image (LDI) approach introduced by Shade et al. [106]) extends a simple depth image to contain several layers of information, having several depth values for each captured pixel. An extension by Chang et.al, the LDI Tree [26] addresses the sampling rate issue to have faster rendering results. These approaches are well suited for depth sensor scenarios, and properly address the occlusion problem with depth images, as seen on figure 4 that depth images cannot represent those. It has been used on VBR scenarios such as works from Daribo and saito [35], Kirshantan et. al [64] Muller et al. [90] and Yoon et al. [125] as Layered Depth Videos, a video extension to LDIs. Significant decreases on data storage with minimized loss of information have been achieved on the tested scenarios when compared to more conservative representation; multiview plus depth (MVD) videos [87]. Differently than LDV's these do not combine and warp information to a single point of view, but transmits each independent point of view as a separate stream.

Surflets [25] have been used to address the same issue as the LDIs addressed. For each captured pixel we may have more than one depth value, and this can be described as a multidimensional function, to which surflets fit as a descriptor.

The work from Zhang and Chen [129] surveys the most common representations applied on IBR applications. This work considers the classical plenoptic function representation of a visual system as the most complete description, and the one that is intended for most IBR applications, hence classifying the alternatives always with the possibility of reconstructing the plenoptic function from the data stored. Although this is not always necessary since there are alternative ways to perform Image-based rendering, this review stays relevant as a survey of common representations. The survey from Smolic et al. [108] also provides an overview on the more recent formats used for Free-viewpoint video, and is an interesting insight on more recent trends.

### 7.1 Summary and Comparison

All these surveyed representations are efficient and fill their purpose on specific types of applications. While meshes are the best representation for human created content due to its ease of manipulation, point clouds are more fitting for image base systems, since a relation between an image pixel and a world coordinate point is easier to achieve than extracting geometric primitives. These representations are geometrically complete, but their higher complexity may render them inefficient on video scenarios.

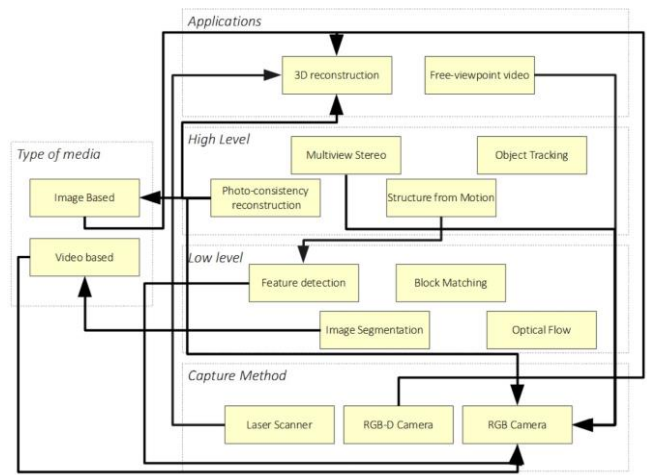

**Fig. 5** Figure highlighting the most relevant association between topics. A connection means that the keywords were both found in more than 70% of the reviewed works

Visual hulls and Image-based representations are geometrically incomplete, but more lightweight on scenarios with big amounts of data .Surflets can also be a more complete description than plain images and show up as a good alternative for the classical representations. The LDI and its derivatives are an interesting upgrade to the depth image representation, and can hold a higher amount of detail than image sets, bringing them closer to the point cloud representation. Having already been applied to VBR scenarios, they make a strong argument as one of the most fitted to such systems.

Of the presented alternative, we believe the image-based representations are more suited for the task since similar approaches from 2D video compression can be adapted to this scenario, as seen on the several works applying the MVD or LDV representations.

## 8 Conclusion and classification

As VBR is a somewhat recent field of research, several of the reviewed works were not directly related to the process, but relevant examples of techniques that play a big part on the pipeline. Classifying these approaches using a hierarchical structure is also infeasible, given the fact that video-based rendering is such a high level definition that largely different applications might fall under the given definition on section 2. Table 1 shows the classification we believe is the most fitting for VBR and what guided our work. For each topic relevant references are outlined, with several references appearing across different topics, since they put different techniques to use.

Multiview Stereo is normally seen as a process to be performed on color cameras (90% of the MVS reviewed articles, e.g. [20] [11] [131]) but, a Laser scanner or color-depth camera setup can also use MVS to match and align its views, if one chooses to do it on image space [120]. We could verify a very common coupling of 3D reconstruction, camera techniques, and MVS, being the most popular option for image-based approaches. [118] [39] [47] [41].

| Survey topic | References |
| --- | --- |
| Block matching | [56][122] |
| Change detection | [99] |
| Segmentation | [22][18][93] |
| Optical flow | [95][7] |
| Features | [19] |
| Multiview stereo | [105] |
| Object detection/tracking | [98][113][124] |
| Image-based rendering | [107] |
| Free viewpoint TV | [128] |
| Representation | [66][129] [108] |
| Video-based rendering | [17][111][114][84] |

**Table 2** List of relevant surveys about the covered topics on this paper

Figure 5 shows connections between keywords that were verified in more than 70% of works reviewed on that category. Some things could be verified such as image segmentation being usually necessary when a VBR process has a main point of interaction/focus, making it an essential step for not only 3D reconstruction applications [76] [80][24] but also free viewpoint video, [52] [63]. This is the lower level technique most cited on video-based works. Although features are used extensively, they were less crucial on most VBR works, being an important step for high level techniques such as SfM (4 out of 5 works).

Although optical flow and block matching are techniques specifically tailored for video, they have not been applied on VBR processes extensively. The work from Pons et al. [96] and Yang et al.[123] are relevant examples of success that should be followed by future work developing on this area.

Feature detection and Multiview Stereo are by far the most standard and widely applied techniques due to the typical necessity of a multi-stream capture setup. Both of these techniques aim to solve this particular problem, each one of them on a different level. A good example of their flexibility and widespread use are the SIFT features [81]. Specific extensions for meshes [83], videos [103], poorly lit scenarios [115] and other examples have been developed. The huge development on these areas also come from the crossover these techniques have with image-based rendering, an older and more developed field.

There is a tendency of increasing attention to Laser Scanner setups, similarly to what happened with color-depth cameras in 2010 with the release of Microsoft Kinect. The more accessible price for these high quality devices will most likely propel the development of 3D reconstruction applications using not only still images, but also three-dimensional video data. Lower level techniques such as optical flow and block matching, and also object tracking and structure from motion, techniques that are tailored for continuous data, will surely play a big role if this type of media develop.

Finally, table 2 list relevant surveys about every approach topic. We recommend those works as more detailed investigation on each one of the covered topics by our survey, and also previous VBR surveys for more theoretical groundwork on the field [84] or a different take on the same process [111] [132].

| Topic / Year | 1980-1989 | 1990-1999 | 2000-2005 | 2006-2010 | 2010-2015 |
|---|---|---|---|---|---|
| Camera input (Section 3) | | [92] [43] [40] [104] [112] | [70] [86] [61] [68] [24] [41] [96] | [47] [63] [126] [131] [97] | [8] [42] [44] [80] [110] [1] [31] [52] [117] [11] [39] [116] [118] [20] [71] [76] [57] [78] [123] [38] |
| Laser Scanner (Section 3) | [16] | [59][6] | [74] | | [110] [54] [55] [69] |
| Depth Sensor (Section 3) | | [34] | | [72] | [88] [58] [73] [39] [5] [23] [119] [76] [120] [127] [123] [38] |
| Optical flow (Section 4.4) | [53] [82] | [46] [9] [77] | [37] [96] [21] | [29] | [48] |
| Block matching (Section 4.1) | [67] | | [130] | [29] | [48] [32] [33][38] |
| Features (Section 4.3) | [82] [51] | [92] [79] | [61] [29] [81] [96] | [72] [103] [10] [8] [44] [83] | [1] [31] [14] [11] [115] |
| Segmentation (Section 4.2) | [12] | [30] | [24] [45] [100] [85] [23] | [8] [110] [80] [60] [91] [63] [131] | [76] [11] [3] [52] [4] |
| Photo-consistency reconstruction (Section 5.1) | | | [70] [86] | | [71] [118] [42] |
| Structure from Motion (Section 5.2) | | | [29] | [109] | [1] [31] [8] |
| Object Recognition (Section 5.3) | | | [86] | [8] | [101] |
| Multiview Stereo (Section 5.1) | | [40] [104] [43] [92] | [96] [41] [68] [61] | [44] [63] [131] [47] | [120] [20] [48] [118] [11] [39] [116] [52] [117] |
| 3D Reconstruction (Section 6.2) | | [92] [43] [34] [40] [104] | [86][70] [68] [24] [41] | [47] [109] [72] [97] [126] [44] [42] [110] [80] | [117] [1] [58] [73] [118] [11] [39] [119] [120] [20] [71] [76] [76] [127] [54] [55] [69] [57] [48] |
| Free viewpoint television 3D Reconstruction (Section 6.1) | | | [61] | [8] [63] | [78] [48] [116] [52] [123] [38] |
| Image-Based techniques | | [40] [43] [92] | [41] [68] [86] [70] | [44] [42] [126] [131] [47] [109] [72] | [69] [20] [71] [76] [55] [118] [11] [39] [119] [73] |
| Video-Based techniques | | | [96] [24] [100] [61] | [110] [80] [63] [103] | [57] [101] [48] [76] [120] [127] [8] [54] [116] [52] [117][1] [58][123][38] |
| Representation (Section 7) | | [26] [106] [28] | [2] [94] | [25] [62] [90] [125] | [64] [35] |

**Table 1** Reviewed works on this survey grouped by reffered techniques and publishing period.